\documentclass[twoside,11pt,preprint]{article}

\usepackage{blindtext}

\usepackage{dmlr2e}

\setcitestyle{authoryear}
\usepackage{pdflscape}
\usepackage{tablefootnote}
\usepackage[utf8]{inputenc} 
\usepackage[T1]{fontenc}    
\usepackage{float}
\usepackage{url}            
\usepackage{booktabs}       
\usepackage{amsfonts}       
\usepackage{amsmath}
\usepackage{nicefrac}       
\usepackage{microtype}      
\usepackage{graphicx}
\graphicspath{ {./figures/} }
\usepackage{adjustbox}
\usepackage{xcolor}
\definecolor{bluecite}{HTML}{0875b7}
\hypersetup{
  colorlinks=true,
  citecolor=bluecite,
  linkcolor=magenta
}

\usepackage[compatibility=false]{caption}
\usepackage{subcaption}
\usepackage{multirow}
\usepackage{bm}

\usepackage{pifont}
\definecolor{cmarkcolour}{HTML}{219e1c}
\definecolor{xmarkcolour}{HTML}{c4231a}

\definecolor{plotblue}{HTML}{5799C7}
\definecolor{plotorange}{HTML}{FF9F4A}
\definecolor{plotgreen}{HTML}{61B861}
\definecolor{plotred}{HTML}{E05D5E}
\definecolor{plotpurple}{HTML}{AF8CCD}

\newcommand{\dotblue}{\color{plotblue}{\Large $\bullet$}}%
\newcommand{\dotorange}{\color{plotorange}{\Large $\bullet$}}%
\newcommand{\dotgreen}{\color{plotgreen}{\Large $\bullet$}}%
\newcommand{\dotred}{\color{plotred}{\Large $\bullet$}}%
\newcommand{\dotpurple}{\color{plotpurple}{\Large $\bullet$}}%

\usepackage{setspace}

\usepackage[nameinlink]{cleveref}

\usepackage{tcolorbox}
\newtcolorbox{mybox}[2]{colback=bluecite!5!white,colframe=bluecite!75!black,fonttitle=\bfseries,title=#1,label=#2}

\usepackage{lastpage}
\dmlrheading{24}{2024}{1-\pageref{LastPage}}{9/23; Revised 10/23}{11/23}{21-0000}{Formanek et al} 

\ShortHeadings{ }{ }
\firstpageno{1}

\begin{document}

\title{Putting Data at the Centre of\\Offline Multi-Agent Reinforcement Learning}

\author{\name Claude Formanek \email{c.formanek@instadeep.com} \\
       \addr University of Cape Town \& InstaDeep, South Africa
       \AND
       \name Louise Beyers \email{l.beyers@instadeep.com} \\
       \addr InstaDeep, South Africa
       \AND
       \name Callum Rhys Tilbury \email{c.tilbury@instadeep.com} \\
       \addr InstaDeep, South Africa
       \AND
       \name Jonathan P. Shock \email{jonathan.shock@uct.ac.za} \\
       \addr University of Cape Town, South Africa
       \AND
       \name Arnu Pretorius \email{a.pretorius@instadeep.com} \\
       \addr InstaDeep, Rwanda
}

\editor{My editor}

\maketitle

\begin{abstract}%
Offline multi-agent reinforcement learning (MARL) is an exciting direction of research that uses static datasets to find optimal control policies for multi-agent systems. Though the field is by definition data-driven, efforts have thus far neglected data in their drive to achieve state-of-the-art results. We first substantiate this claim by surveying the literature, showing how the majority of works generate their own datasets without consistent methodology and provide sparse information about the characteristics of these datasets. We then show why neglecting the nature of the data is problematic, through salient examples of how tightly algorithmic performance is coupled to the dataset used, necessitating a common foundation for experiments in the field. In response, we take a big step towards improving data usage and data awareness in offline MARL, with three key contributions: (1) a clear guideline for generating novel datasets; (2) a standardisation of over 80 existing datasets, hosted in a publicly available repository, using a consistent storage format and easy-to-use API; and (3) a suite of analysis tools that allow us to understand these datasets better, aiding further development. These contributions are all publicly available on our website.\footnote{\url{https://instadeepai.github.io/og-marl/}}
\end{abstract}


\section{Introduction} \label{sec:intro}
Many complex real-world problems can naturally be formulated as multi-agent systems---e.g. managing traffic~\citep{zhang2019cityflow}, controlling fleets of ride-sharing vehicles~\citep{sykora2020marvin} or a network of trains~\citep{mohanty2020flatland}, optimising electricity grid usage~\citep{khattar2022citylearn}, and improving dynamic packet routing in satellite communication~\citep{lozano2024open}. Improving on solutions to such problems is an important endeavour, because of the potentially immense societal benefits that they offer. Multi-Agent Reinforcement Learning (MARL) is a promising avenue to finding solutions to such problems, but the field faces a host of hurdles which must first be overcome. One key difficulty is the access to accurate and efficient simulators, for online experience generation and exploration. To learn robust policies, extensive interactions with an environment are usually required~\citep{yu2018towards}, which makes simulator efficiency of paramount importance. Yet, for real-world applicability, the fidelity between the simulator and reality must also be maintained.  Unfortunately, this balance of achieving high throughput in an online simulator while maintaining realistic dynamics is difficult, and practitioners must often rely on more basic environments with simplifying assumptions. The situation is particularly challenging when there are many agents interacting in complex ways, as is the case in MARL.

What typically does exist in such systems as those described above is the ability to capture large amounts of useful data. Across a range of complex control scenarios, even in situations where many agents are acting and the physical dynamics are not well understood (i.e. where designing a bespoke simulator would be very challenging), it may be straightforward to record data during operation. This opportunity is what \emph{offline} RL leverages, by bridging the gap between RL and supervised learning. In the offline domain, the aim is to develop algorithms that use large, existing datasets of sequential decision-making transitions (whether recorded from the real-world, or created in simulation, or a mixture thereof) to learn optimal control policies, which can later be deployed online~\citep{levine2020offline}. The offline paradigm promises to help unlock the full potential of RL when applied to the real-world, where success has thus far been limited~\citep{dulac2021challenges,fig_rlreview}. In the multi-agent setting, algorithms are designed to learn a \textit{joint} policy from a static dataset of previously collected multi-agent transitions, generated by a set of interacting behaviour policies.

Single-agent offline RL has enjoyed relatively widespread research attention and success~\citep{prudencio2023offlinerl}. Core to such developments, the community has benefited greatly from standardised and publicly available datasets---as found in libraries such as D4RL~\citep{fu2020d4rl} and RL Unplugged~\citep{gulcehre2020rl}. Yet, in the multi-agent case, such offerings are limited both in number and in quality. In fact, we argue in this paper that work in offline MARL has disproportionately focused on algorithmic innovation, and neglected the role of data almost entirely. It is common practice for authors to generate their own datasets for their experiments, almost as an after-thought, while little effort has been made to understand how the quality and content of these multi-agent datasets affect training dynamics and final performance. Such insights have been immensely valuable in the single-agent setting~\citep{schweighofer2022dataset}, and it is known that multi-agent systems face additional complexities~\citep{tilbury2024coordination}, where similar insights would be particularly useful.

Importantly, it should be clear that the end-goal for offline MARL is for systems to be deployed in the real world: real datasets, yielding real policies, useful in real applications. Yet, the path there is long and winding. To make progress, we focus in this paper on a simplified context: cooperative scenarios (where existing online MARL research is most mature, and many real-world examples exist), with data recorded from simulators. Solving these simpler problems does not magically solve the more complex ones, but we start here as a necessary first step towards deploying MARL in the wild. Notice that this journey parallels the one that computer vision has undertaken, as an example of a more established field of machine learning---before starting to tackle the enormously challenging task of, say, end-to-end learning from raw pixels for a self-driving car in the real-world (e.g.~\citet{bojarski2016end}), it was important to first tackle simpler problems, like handwritten digit recognition~\citep{lecun1989backpropagation}. Focusing on fundamental datasets like MNIST~\citep{lecun2010mnist} has been integral to the progress in computer vision. We make similar efforts for offline MARL here.

Our paper is structured in the following way. We start surveying the current state of the field in Section~\ref{sec:state-of-the-field}, by studying how authors have, until now, been treating data in their research---with the evidence showing a general lack of care in the way data is considered. We build on this finding in Section~\ref{sec:impact-of-data} to show why this carelessness is problematic. Through four clear examples, we show how the specifics of the data has a significant impact in the learned performance of algorithms, something which has previously been overlooked. We respond in Section~\ref{sec:solutions} with three contributions: firstly, a clear guideline on how data should be treated in offline MARL going forward; secondly, a standardised set of datasets, comprising over 80 \texttt{environment-scenario-quality} combinations, with a well-documented and accessible API, and an easy mechanism to extend this repository; and finally, useful tooling for researchers to understand the nature of their datasets, as an initial effort to promote data awareness in offline MARL.

\section{The Current State of Datasets for Offline MARL} \label{sec:state-of-the-field}
Offline MARL remains a relatively nascent field, with only a handful of papers released on the topic to date (see Table \ref{tab:datasets_summary}), but progress is accelerating. We want to understand how authors have been handling the data component of their research in the work done thus far by trying to answer the question: what is the state of the field, with respect to data itself? To do so, we present a comprehensive survey of work in empirical offline MARL, from leading peer-reviewed academic venues, to assess (1) how their data were generated and (2) what information about their datasets was provided. Though a simple assessment, we find these two axes already particularly telling in what they reveal. Table~\ref{tab:datasets_summary} summarises our findings.

\begin{table}[p]
\centering
\scriptsize
\setstretch{1.8}
\caption{A survey of the use of data in empirical offline MARL literature, focusing on (1.) where the data came from, and (2.) which properties of the data were reported. We notice that most of the assessed papers self-generated their data. We also see that there is no consistent naming strategy in the dataset qualities, and that the mean and standard deviation of the datasets are often absent.} \label{tab:datasets_summary}
\begin{adjustbox}{max width=\textwidth}
\begin{tabular}{c|c|ccc}
\hline
\textbf{Paper} & \textbf{Source of Datasets} & \textbf{Dataset Labels} & \textbf{Means}, $\mu$ & \textbf{Standard Deviations} \\
\hline

\begin{tabular}{c}
MAICQ \\ \citep{maicq}
\end{tabular}
&
\begin{tabular}{c}
\textcolor{xmarkcolour}{\textbf{Self-generated}} using \\ DOP~\citep{wang2021dop}
\end{tabular}
& \begin{tabular}{c} 
\emph{good} \\
\emph{medium} \\
\emph{poor}
\end{tabular} & \begin{tabular}{c}
$15<\mu<20$ \\
$10<\mu<15$ \\
$0<\mu<10$
\end{tabular} &
\textcolor{xmarkcolour}{\textbf{Not given}}
\\ \hline

\begin{tabular}{c}
MADT \\ \citep{madt}
\end{tabular}
& 
\begin{tabular}{c}
\textcolor{xmarkcolour}{\textbf{Self-generated}} using \\ MAPPO~\citep{yu2022surprising}
\end{tabular}
& \begin{tabular}{c} \emph{replay} \end{tabular} & Given & Given
\\ \hline

\begin{tabular}{c}
Offline MARL with \\ Knowledge distillation \\ \citep{tseng2022knowledgedistill}
\end{tabular}
&
\begin{tabular}{c}
Some \textcolor{xmarkcolour}{\textbf{self-generated}} using random policies or\\expert policies from PPO~\citep{schulman2017proximal},\\and some from MADT~\citep{madt}
\end{tabular}
& \begin{tabular}{c} 
\emph{good} \\
\emph{normal} \\
\emph{poor}
\end{tabular} & Given  & Given
\\ \hline

\begin{tabular}{c}
OMAR \\ \citep{omar}
\end{tabular}
&
\begin{tabular}{c}
\textcolor{xmarkcolour}{\textbf{Self-generated}} using \\ MATD3~\citep{ackermann2019matd3}
\end{tabular}
& \begin{tabular}{c} 
\emph{random} \\
\emph{medium-replay} \\
\emph{medium} \\
\emph{expert} \\
\end{tabular} &
\textcolor{xmarkcolour}{\textbf{Not given}} &
\textcolor{xmarkcolour}{\textbf{Not given}}
\\ \hline

\begin{tabular}{c}
CFCQL \\ \citep{cfcql}
\end{tabular}
&
\begin{tabular}{c}
\textcolor{xmarkcolour}{\textbf{Self-generated}} using \\ QMIX~\citep{rashid2018qmix}
\end{tabular}
& \begin{tabular}{c}
\emph{medium} \\
\emph{medium-replay} \\
\emph{expert} \\
\emph{mixed}
\end{tabular} &
\textcolor{xmarkcolour}{\textbf{Not given}} &
\textcolor{xmarkcolour}{\textbf{Not given}}
\\ \hline

\begin{tabular}{c}
OMAC \\ \citep{omac}
\end{tabular}
& \begin{tabular}{c}
Random Subsets from\\MADT~\citep{madt}
\end{tabular}
& \begin{tabular}{c} 
\emph{good} \\
\emph{medium} \\
\emph{poor}
\end{tabular}
& \textcolor{xmarkcolour}{\textbf{Not given}} & \textcolor{xmarkcolour}{\textbf{Not given}}
\\ \hline

\begin{tabular}{c}
OMIGA \\ \citep{omiga}
\end{tabular}
& \begin{tabular}{c}
Random Subsets from\\MADT~\citep{madt}
\end{tabular}
& \begin{tabular}{c} 
\emph{good} \\
\emph{medium} \\
\emph{poor}
\end{tabular}
& Given & \textcolor{xmarkcolour}{\textbf{Not given}}
\\ \hline

\begin{tabular}{c}
SIT Framework \\ \citep{tian2023goodtrajectories}
\end{tabular}
&
\begin{tabular}{c}
\textcolor{xmarkcolour}{\textbf{Self-generated}} using \\ QMIX~\citep{rashid2018qmix}\\ and FacMAC~\citep{peng2021facmac}
\end{tabular}
& \begin{tabular}{c} 
\emph{low quality} \\
\emph{medium quality} \\
\emph{random quality} \\
\end{tabular} & Given  & \textcolor{xmarkcolour}{\textbf{Not given}} 
\\ \hline

\begin{tabular}{c}
AlberDICE \\ \citep{alberdice}
\end{tabular}
&
\begin{tabular}{c}
\textcolor{xmarkcolour}{\textbf{Self-generated}} using\\MAT~\citep{wen2022mat}
\end{tabular}
& \begin{tabular}{c} 
\emph{expert} \\
\emph{medium-expert} \\
\end{tabular} &  \textcolor{xmarkcolour}{\textbf{Not given}}   & \textcolor{xmarkcolour}{\textbf{Not given}}
\\ \hline

\begin{tabular}{c}
Value Deviation \&\\Transition Normalisation\\\citep{jiang2023offline}
\end{tabular}
&
\begin{tabular}{c}
Some \textcolor{xmarkcolour}{\textbf{self-generated}} using random policies\\or using QMIX~\citep{rashid2018qmix}\\or SAC~\citep{sac},\\and some decomposed from D4RL~\citep{fu2020d4rl}
\end{tabular}
& \begin{tabular}{c} 
\emph{random} \\
\emph{medium} \\
\emph{replay} \\
\emph{expert} \\
\end{tabular} &  \textcolor{xmarkcolour}{\textbf{Not given}}   & \textcolor{xmarkcolour}{\textbf{Not given}} 
\\ \hline

\begin{tabular}{c}
State Augmentation via Self-Supervision\\\citep{wang2023state}
\end{tabular}
&
\begin{tabular}{c}
\textcolor{xmarkcolour}{\textbf{Self-generated}} using\\QMIX~\citep{rashid2018qmix}
\end{tabular}
& \begin{tabular}{c} 
\emph{inadequate} \\
\emph{moderate} \\
\emph{superb} \\
\end{tabular} & 
\begin{tabular}{c}
$0<\mu<10$ \\
$10<\mu<15$ \\
$15<\mu<20$
\end{tabular}
& \textcolor{xmarkcolour}{\textbf{Not given}} 
\\ \hline

\begin{tabular}{c}
MADiff\\\citep{zhu2023madiff}
\end{tabular}
&
\begin{tabular}{c}
Datasets from~\citet{formanek2023ogmarl}\\and from~\citet{omar}
\end{tabular}
& \begin{tabular}{c} 
\emph{good} \\
\emph{medium} \\
\emph{poor} \\
\end{tabular} & Given & Given 
\\ \hline

\begin{tabular}{c}
MOMA-PPO \\ \citep{barde2024model}
\end{tabular}
&
\begin{tabular}{c}
\textcolor{xmarkcolour}{\textbf{Self-generated}} using \\ MAPPO~\citep{yu2022surprising},\\ and decomposed from D4RL~\citep{fu2020d4rl}
\end{tabular}
& \begin{tabular}{c} 
\emph{random} \\
\emph{medium} \\
\emph{replay} \\
\emph{expert} \\
\emph{expert-mix}
\end{tabular} & Given  & \textcolor{xmarkcolour}{\textbf{Not given}} 
\\ \hline
\end{tabular}
\end{adjustbox}
\end{table}

From Table~\ref{tab:datasets_summary}, we firstly notice that the majority of papers assessed generated their own datasets. Each paper also creates these datasets in different ways---using a wide variety of underlying online algorithms to learn policies for the generated trajectories. The dataset labels themselves also vary across papers, with no consistent naming convention. Information about the dataset properties is also sparse. To measure a given dataset's `quality', the mean episode return of trajectories is often reported, but even this metric is not always given. The return distribution is mostly ignored, except for the occasional proxy of reporting the standard deviation. Essentially, there is little information presented on the contents and diversity of the experience in a given dataset. Yet these aspects of a dataset are crucial to the resulting performance of offline learning. In the single-agent literature, it has been shown that dataset properties have a marked impact on results~\citep{schweighofer2022dataset}. In the multi-agent context, other complex aspects of coordination~\citep{tilbury2024coordination, barde2024model} make the contents and characteristics of datasets even more important to understand.

We observe here that the field of offline MARL has struggled to find common ground to benchmark proposed algorithms. Even accepting that many authors generate their own datasets for their papers, there has been carelessness in reporting information about such datasets. Ultimately, the current reality points to a general lack of consideration around the role of data in the field, where authors are failing to adequately control for the impact that data can have on experimental results. We will now show why a lack of such data-centrism is problematic for the field.

\section{Why Dataset Characteristics Matter}\label{sec:impact-of-data}

\begin{figure}
    \centering
    \begin{tabular}{c}
        \begin{subfigure}[p]{0.35\textwidth}
            \centering
            \scriptsize
            \setstretch{1.5}
                \begin{tabular}{c|c}
                & $\mu$ \\\hline
                \dotred & $7$  \\
                \dotgreen & $10$  \\
                \dotorange & $13$  \\
                \dotblue & $16$  \\
                \bottomrule
                \end{tabular}
        \caption{Mean episode return for each dataset. Datasets from the three SMAC scenarios \texttt{5m\_vs\_6m}, \texttt{3s5z\_vs\_3s6z} and \texttt{2s3z} were generated for each mean episode return.}
        \label{fig:mean-matters-means}
        \vspace{1em}
      \end{subfigure}\\
      \end{tabular}
    \hspace{1em}
    \begin{subfigure}[p]{0.55\textwidth}
         \centering
         \includegraphics[width=\textwidth]{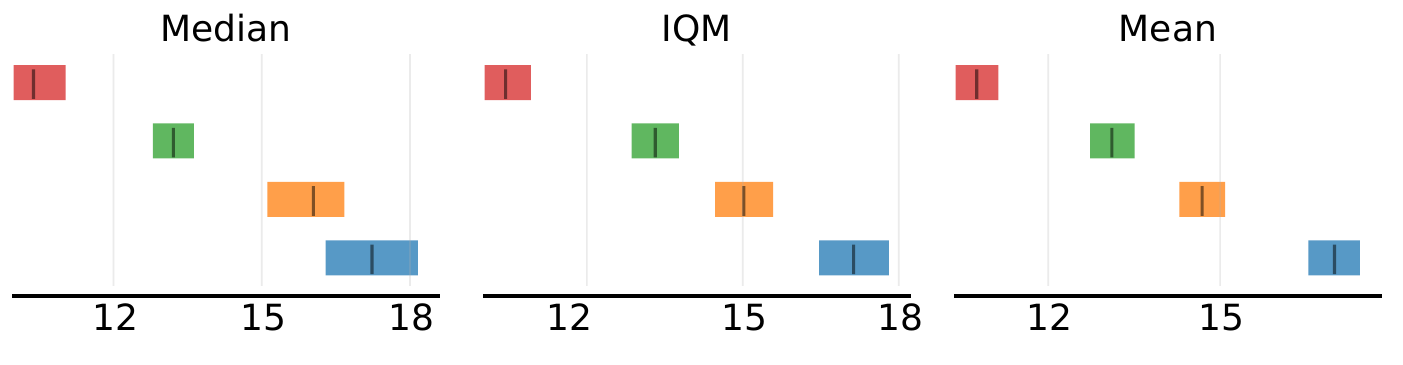}
         \caption{A comparison of the performance obtained using datasets with different mean episode returns. Results are aggregated (across two algorithms, 10 random seeds, 32 evaluation episodes and three scenarios) using bootstrap confidence intervals \citep{agarwal2021deep}.}
         \label{fig:mean-matters-results}
    \end{subfigure}
      \caption{To demonstrate the effect that the mean episode return of a dataset has on the final performance of an offline MARL algorithm, we generated four datasets with mean episode returns given in \autoref{fig:mean-matters-means}. We then train an offline MARL algorithm for 50k training steps on each of the datasets and compare the final performance of the algorithm across the different datasets. We repeat the experiment across three different SMAC scenarios, two different algorithms (\texttt{IQL+CQL}~\citep{formanek2024dispelling} and \texttt{MAICQ}~\citep{maicq}) and 10 random seeds. The aggregated results are given in \autoref{fig:mean-matters-results}.}
\label{fig:mean-matters}
\end{figure}

Claims of algorithmic improvement become moot if acommon basis of data is missing, since a dataset is one of the control variables in empirical offline MARL experiments which can impact performance significantly. We illustrate this point by giving four examples that progressively show how algorithmic results are tightly coupled with data. These examples are not intended to make sweeping claims about the field, but should rather serve as a series of ``proof by existence'' demonstrations cautioning researchers of how peculiarities of datasets may be influencing their experimental findings. When reflecting on this evidence, it becomes clear that overlooking data is problematic for the field, and that ultimately, there is a serious need for a shift in current research practices.

\paragraph{\textbf{Dataset Mean.\;}} We begin our illustration with likely the most intuitive example: what happens to final performance when the average return of the dataset changes? We construct four distinct datasets with increasing means on three scenarios (\texttt{5m\_vs\_6m}, \texttt{3s5z\_vs\_3s6z} and \texttt{2s3z}) from the SMACv1 environment~\citep{samvelyan2019smac}, by subsampling episodes from OG-MARL datasets~\citep{formanek2023ogmarl}. For each dataset, we fix the standard deviation at approximately 2.0 as calculated over 2000 episodes. We then train two offline MARL algorithms, \texttt{IQL+CQL}~\citep{formanek2024dispelling} and \texttt{MAICQ}~\citep{maicq}, on the individual datasets and report the final evaluation episode return, averaged across 10 random seeds. We present the aggregated results using bootstrap confidence intervals~\citep{agarwal2021deep,gorsane2022emarl} in Figure~\ref{fig:mean-matters}.

We see that the aggregated final performance of \texttt{IQL+CQL} and \texttt{MAICQ} is positively correlated with the mean episode return of the dataset used for training. As the average quality of the data improves, so does the offline learning from this dataset---an intuitive result. This relationship is often used in the single-agent literature, where the final performance is reported as a percentage of the mean return in the dataset~\citep{agarwal2019optimistic, gulcehre2020rl}, but doing so is rare in the multi-agent domain. In fact, the simple metric of the dataset mean is sometimes omitted entirely, as shown in Table~\ref{tab:datasets_summary}.

\begin{figure}
    \centering
    \begin{subfigure}[p]{0.47\textwidth}
         \centering
         \includegraphics[width=\textwidth]{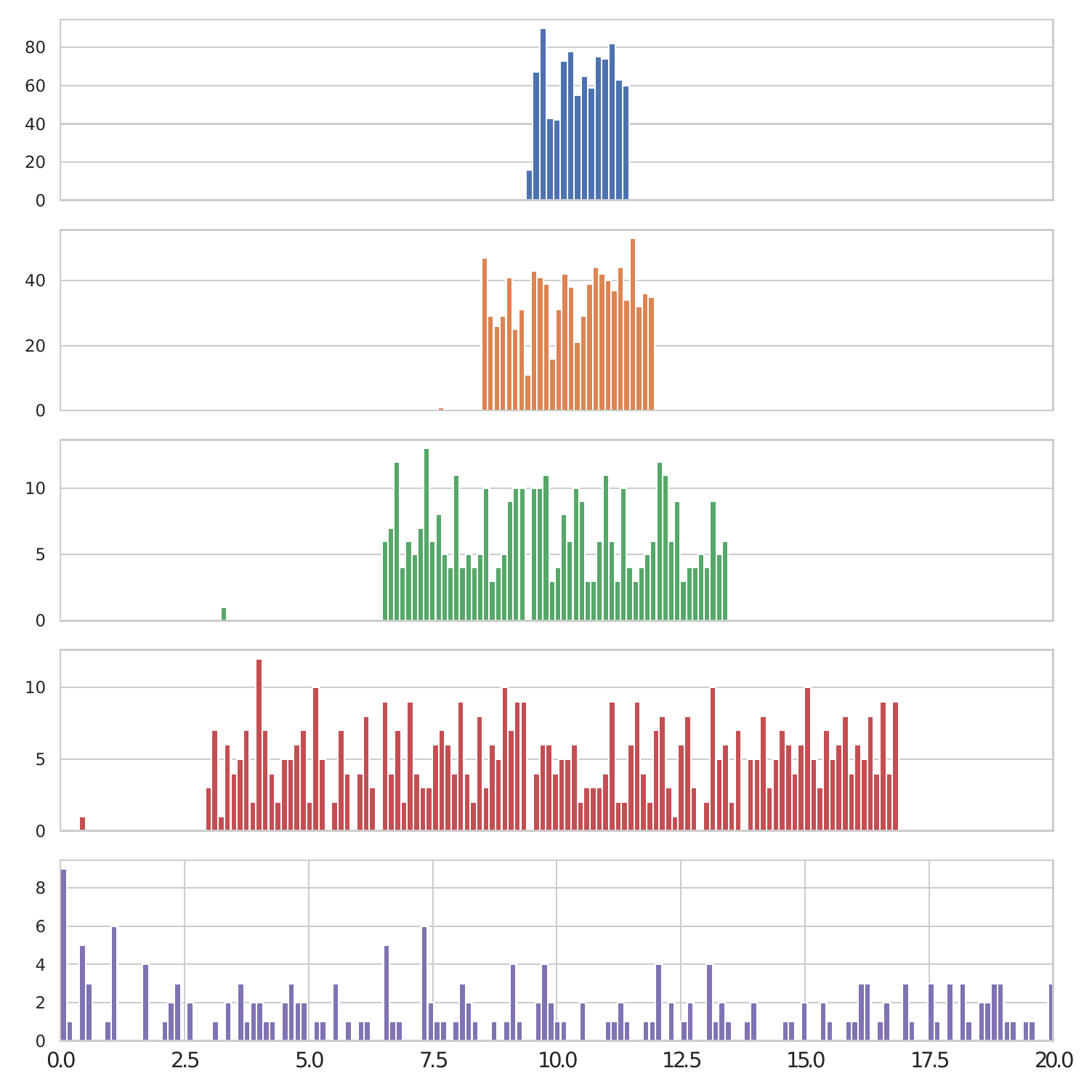}
         \caption{Histograms of the episode returns of five datasets with the same mean but different standard deviations.}
         \label{fig:std_matters_violins}
    \end{subfigure}
    \hspace{1em}
    \begin{tabular}{c}
        \begin{subfigure}[p]{0.46\textwidth}
            \centering
            \scriptsize
            \setstretch{1.5}
                \begin{tabular}{c|cc}
                 & $\mu$ & $\sigma$ \\\hline
                \dotblue & $10$ & $0.5$ \\
                \dotorange & $10$ & $1.0$ \\
                \dotgreen & $10$ & $2.0$ \\
                \dotred & $10$ & $4.0$ \\
                \dotpurple & $10$ & $6.0$ \\
                \bottomrule
                \end{tabular}
        \caption{Dataset mean and standard deviation.}
        \vspace{1em}
      \end{subfigure}\\
      \begin{subfigure}[p]{0.45\textwidth}
         \centering
         \includegraphics[width=\textwidth]{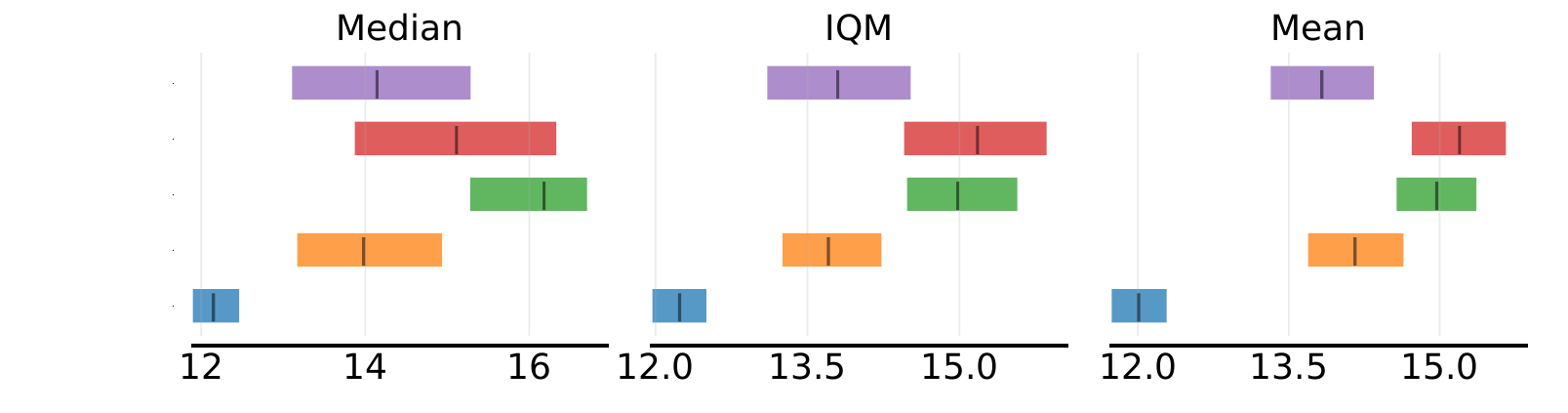}
         \caption{The aggregated results across scenarios (\texttt{5m\_vs\_6m}, \texttt{3s5z\_vs\_3s6z} and \texttt{2s3z}), algorithms (\texttt{IQL+CQL} and \texttt{MAICQ}) and random seeds are given for each dataset. The median, interquartile mean (IQM), and mean are all given with bootstrap confidence intervals \citep{agarwal2021deep}.}
      \end{subfigure}
      \end{tabular}
      \caption{To demonstrate the surprising effect that the standard deviation (std) can have on the performance of an offline MARL experiment we generate 5 datasets that each had the same mean but differing std. We then train two offline MARL algorithms, \texttt{IQL+CQL} and \texttt{MAICQ}, on the data and report the final performance. We repeat the experiment across three different SMAC scenarios, two different algorithms, 10 random seeds and an evaluation batch size of 32. We then aggregate the results as per \cite{gorsane2022emarl}.}
     \label{fig:std_matters}
\end{figure}

\paragraph{\textbf{Dataset Spread.\;}}
Another important property of a dataset in offline RL is the diversity of the experience~\citep{agarwal2019optimistic}. In the single-agent setting, it is well-understood that many offline RL algorithms benefit from more diverse datasets~\citep{schweighofer2022dataset}. Usually, this benefit arises because diverse experience leads to better coverage of the state and action spaces, resulting in fewer out-of-distribution actions, which are known to cause issues when training offline~\citep{fujimoto2019bcq}. However, the effect of diversity in the multi-agent setting is less well understood and should not be taken for granted. For example, \citet{tilbury2024coordination} demonstrate a more complex relationship between dataset diversity and final performance in the offline MARL setting than might be initially assumed. In the literature, the standard deviation (std) of the data's episode returns is sometimes reported as a proxy for diversity.

We thus continue our illustration with another example: what happens when solely the std of the returns in a dataset changes? We now construct five distinct datasets, each with the same mean, but with increasing spread around that mean, keeping the number of included episodes constant. We create these datasets by subsampling from OG-MARL. Once again, we look at the three SMAC scenarios \texttt{5m\_vs\_6m}, \texttt{3s5z\_vs\_3s6z} and \texttt{2s3z}. Figure~\ref{fig:std_matters} shows the respective histograms of the datasets, and the corresponding aggregated results across two algorithms (\texttt{IQL+CQL} and \texttt{MAICQ}) and 10 random seeds.

In this situation, we begin to see a more complex relationship emerge. Rather than a simple linear relationship between diversity and performance, optimal results are found at intermediate levels of std. This result is less intuitive than before (where a higher mean return simply meant higher performance), and begins to hint at the impact of multi-agent dynamics. It is worth reiterating that very few papers report the std (or some other proxy for diversity) of their datasets---just three of the thirteen in Table~\ref{tab:datasets_summary} had done so.

\begin{figure}
    \centering
     \begin{subfigure}[p]{0.46\textwidth}
         \centering
         \includegraphics[width=\textwidth]{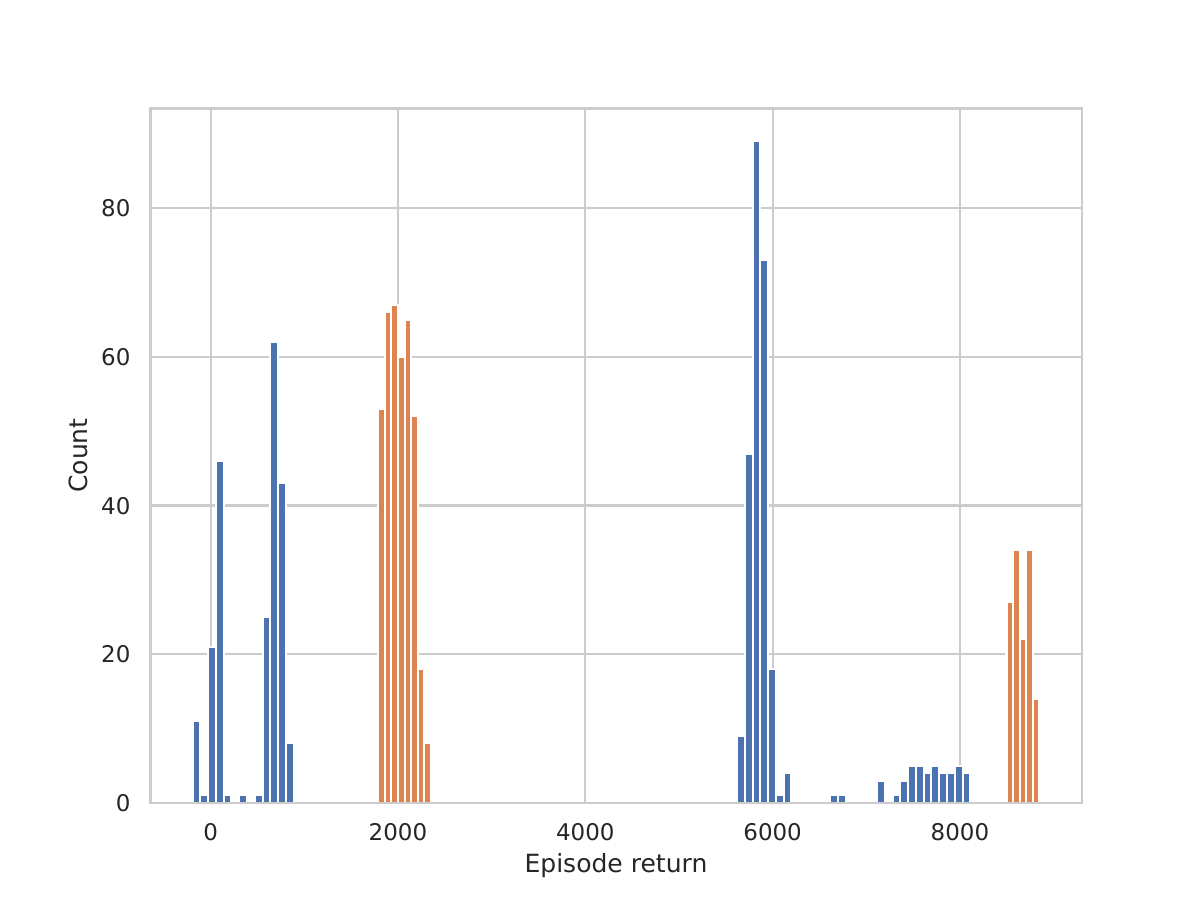}
         \caption{Histogram of the episode returns for two datasets with similar means and standard deviations.}
         \label{fig:same-summary-stats-a}
    \end{subfigure}
    \hspace{1em}
    \begin{tabular}{c}
    \begin{subfigure}[p]{0.46\textwidth}
         \centering
         \scriptsize
         \setstretch{1.5}
         \begin{tabular}{c|cc}
                 & $\mu$ & $\sigma$ \\\hline
                \dotorange & $3688$ & $2885$ \\
                \dotblue & $3665$ & $2866$ \\
                \bottomrule
        \end{tabular} \label{tab:same-summary-stats-b}
        \caption{Mean and standard deviation.}
        \vspace{1em}
      \end{subfigure}\\
    \begin{subfigure}[p]{0.45\textwidth}
        \centering
         \includegraphics[width=\textwidth]{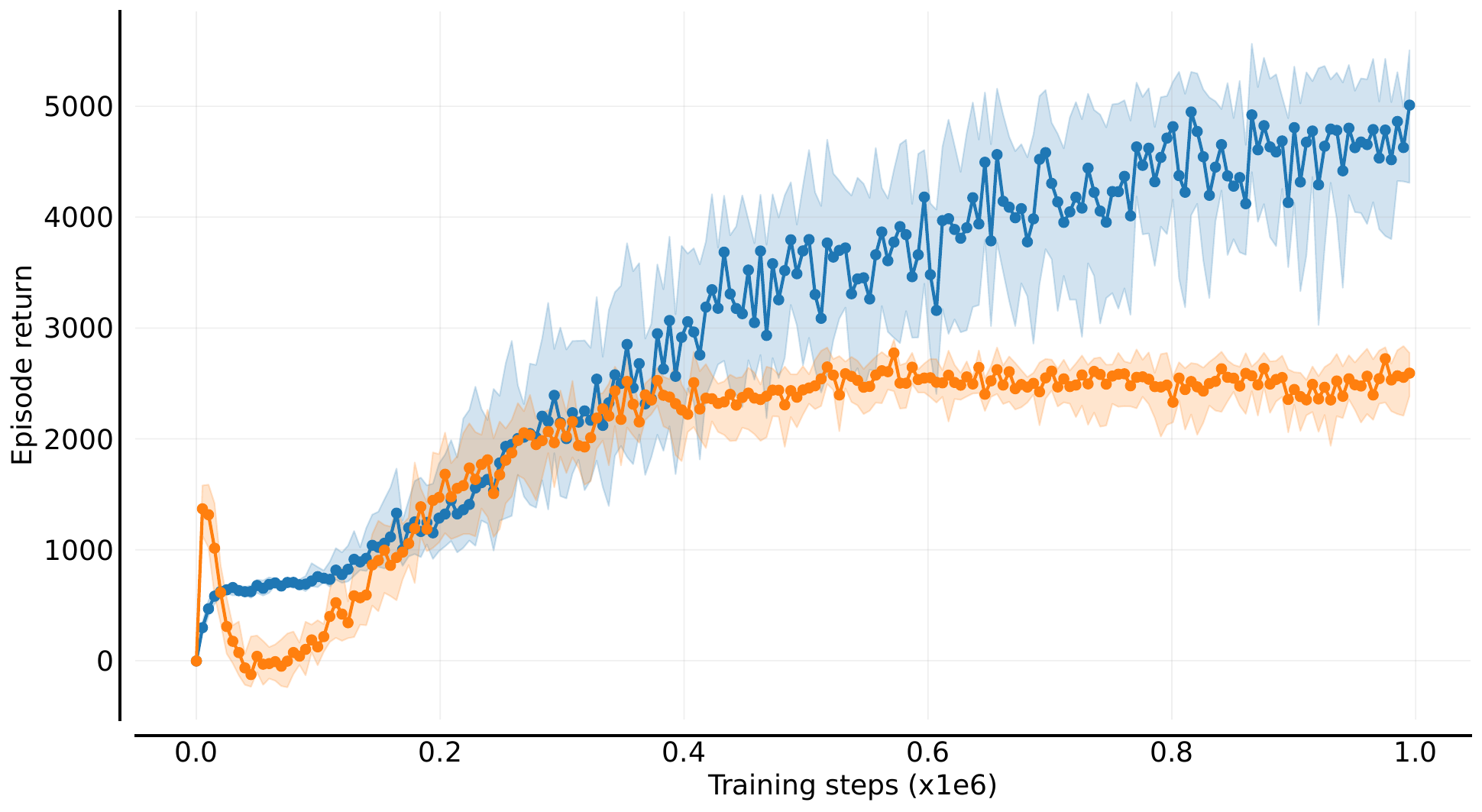}
         \caption{Training curves.
         }
         \label{fig:mean_std_not_enough}
     \end{subfigure} \label{tab:same-summary-stats-c}
    \end{tabular}
     \caption{We generate two datasets on 2-Agent Halfcheetah each with very similar episode return means and standard deviations, but distinct data distributions. We then train \texttt{MADDPG+CQL} on each dataset and report its performance over 1 million training steps. We repeat the experiment over 10 random seeds.}
     \label{fig:same-summary-stats}
\end{figure}

\paragraph{\textbf{Dataset Distribution.\;}}  Our illustration continues with a question that builds on the previous two: controlling for equal mean and std, can two different dataset distributions yield different results? We construct two more datasets subsampled from OG-MARL with this property.\footnote{This setup is reminiscent of \emph{Anscombe's Quartet}~\citep{anscombe1973graphs}, comprising four datasets that have almost identical summary statistics, yet look markedly different when visualised.} We look at the \texttt{2halfcheetah} scenario from MAMuJoCo~\citep{peng2021facmac}. We train \texttt{MADDPG}~\citep{lowe2017maddpg} with \texttt{CQL}~\citep{kumar2020cql} on these respective datasets, and show the results in Figure~\ref{fig:same-summary-stats}.

We can see here that an offline algorithm, when trained on two datasets with nearly identical summary statistics, can yield significantly different final performances. In essence, these metrics of a dataset only paint a limited view of the underlying distributions, yet these distributions may have a notable impact on the results. Here, we note that even those authors from Table~\ref{tab:datasets_summary} who have reported the mean and std of their datasets, have nonetheless omitted a visualisation of their datasets for further understanding. Therefore, authors might be missing key insights on the characteristics of the data they are using in their work.

\paragraph{\textbf{Dataset Coverage.\;}} We conclude our illustration with possibly the most illuminating example of the subtleties of dataset charateristics and their effect on performance. We ask: can we have a situation where the return distributions are very similar, such that the summary statistics are similar and the histograms are closely aligned, yet yield significantly different results for the same offline algorithm?

\begin{figure}
    \centering
    \begin{subfigure}[p]{0.45\textwidth}
            \includegraphics[width=0.8\linewidth]{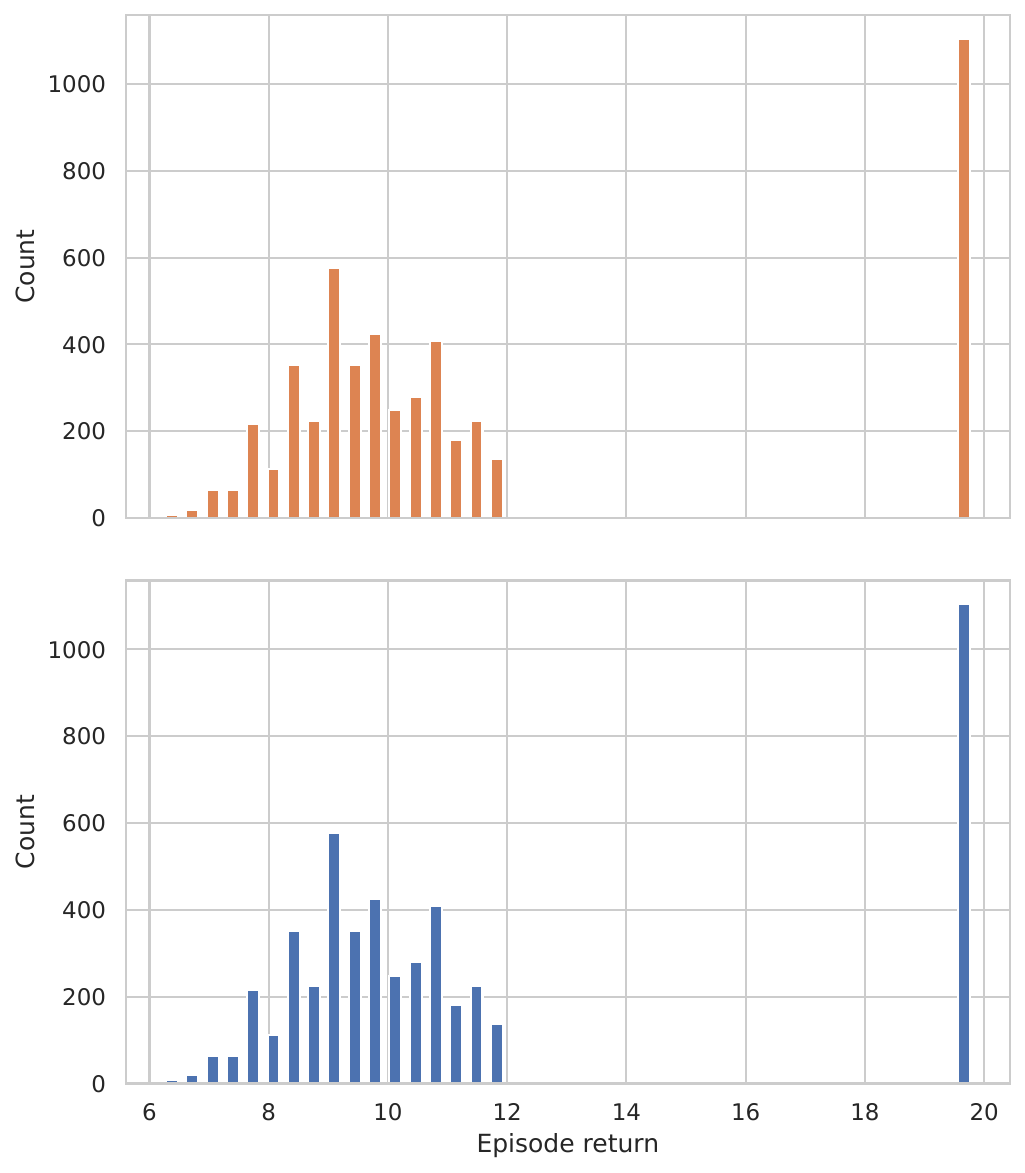}
        \caption{Histograms of the subsampled\\ \texttt{5m\_vs\_6m} datasets.}
        \label{fig:ogmarl-cfcql-violins}
    \end{subfigure}
    \begin{tabular}{c}
    \begin{subfigure}[p]{0.50\textwidth}
         \centering
         \scriptsize
         \setstretch{0.9}
         \caption{Means and standard deviations of the episode returns in subsampled datasets.}
        \begin{tabular}{lcccc}
        \hline
            Dataset & Mean & Stddev & \# Traj & \# Trans \\
            \hline
            \dotorange{} \textcolor{black}{CFCQL} & 12.05 & 4.36 & 4992 & 140073 \\
            \dotblue{} \textcolor{black}{OG-MARL}  & 12.05 & 4.36 & 4992 & 134985 \\
            \hline
            \hline
        \end{tabular} \label{tab:ogmarl-cfcql-summary-stats}
        \vspace{1em}
        \end{subfigure}\\
        \begin{subfigure}[p]{0.45\textwidth}
            \centering
             \includegraphics[width=\textwidth]{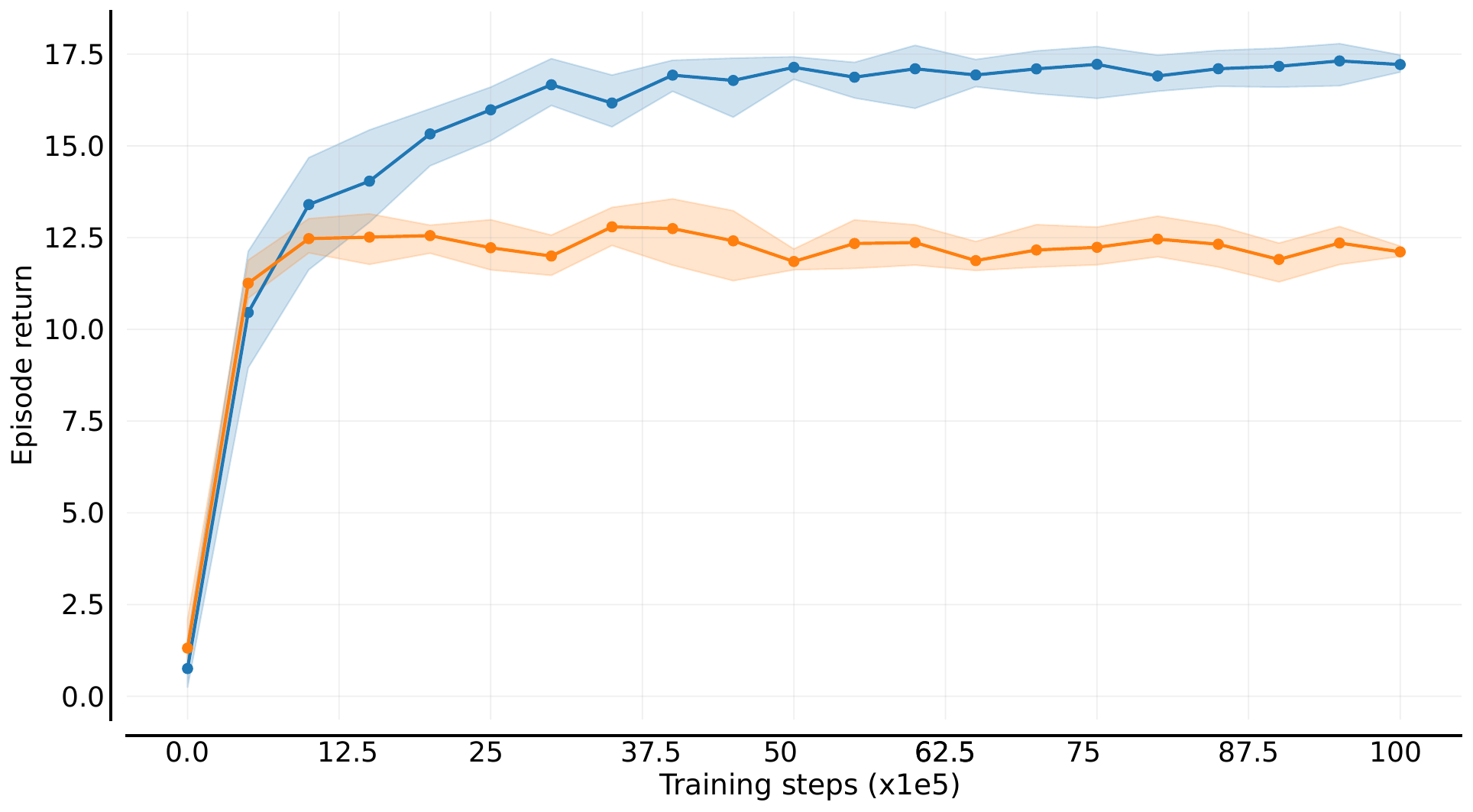}
             \caption{Training curves}
            \label{fig:ogmarl-cfcql-results}
     \end{subfigure} 
    \end{tabular}
    \label{fig:ogmarl-cfcql}
    \caption{We use two subsampled datasets of the \texttt{5m\_vs\_6m} scenario from SMACv1, with almost identical distributions, but from two different sources~\citep{formanek2023ogmarl,cfcql} (the \texttt{Medium} quality in both cases). We then train \texttt{IQL+CQL} on each dataset and report its final performance. We repeat the experiment over 10 random seeds.}
\label{fig:describe_5m_vs_6m}
\end{figure}

For this example, we look at two publicly available datasets, from OG-MARL~\citep{formanek2023ogmarl} and CFCQL~\citep{cfcql} respectively. We consider the \texttt{5m\_vs\_6m} scenario from SMACv1~\citep{samvelyan2019smac}, and take the \texttt{Medium} quality dataset from each source. To further control the experiment, we subsample the original datasets to around $140k$ transitions, matching the distributions of episode returns in the process. The result is not only equal mean and std but also nearly identical, visually indistinguishable histograms, which we show in Figure~\ref{fig:ogmarl-cfcql-violins} and detail in Table~\ref{tab:ogmarl-cfcql-summary-stats}. We train \texttt{IQL+CQL} on both datasets, across 10 random seeds. The results are illustrated in Figure~\ref{fig:ogmarl-cfcql-results}.

\begin{table}[hbtp]
    \centering
    \caption{The Joint State-Action Coverage (Joint-SACo) scores for two datasets on the \texttt{5m\_vs\_6m} SMAC task, one from OG-MARL and the other from CFCQL \citep{cfcql}. A lower score means there is less Joint State-Action coverage in the dataset, i.e. there are more repeated transitions in the dataset.}
    \label{tab:saco}
    \begin{tabular}{llc}
         \hline
         Dataset & Quality & Joint-SACo score \\
         \hline
         \dotorange{} \textcolor{black}{CFCQL} & \texttt{Medium} & 0.10 \\
         \dotblue{} \textcolor{black}{OG-MARL} & \texttt{Medium} & 0.83 \\
         \hline
         \bottomrule
    \end{tabular}
\end{table}

We see here a curious outcome---despite the return distributions being essentially the same, the achieved algorithm performance is significantly different. The difference cannot be explained solely by the statistics and distribution of the episode returns. Evidently, there are other significant differences between the datasets which escape the reach of our current lens. We note that episode return is itself a summary of a trajectory, and is an abstraction of the actual experience in the dataset.

To better understand what is happening here, we extend to the multi-agent setting the approach from \citet{schweighofer2022dataset} to measure dataset diversity. Specifically, we extend their State-Action Coverage (SACo) metric to a version that operates on the joint state and action space of agents in MARL. We define this metric in the same way as SACo: the ratio between the number of unique state-action pairs and the total number of state-action pairs in a dataset. In the multi-agent case, the difference is that the action refers to the joint action of all agents in the system. We use this Joint-SACo metric on the \texttt{5m\_vs\_6m} datasets from Figure~\ref{fig:ogmarl-cfcql-violins}, with the results given in Table \ref{tab:saco}.

Whereas the datasets from this example have almost identical return distributions, we see that they have very different values for state-action coverage. In fact, from Table \ref{tab:saco} we observe that $90\%$ of the data from~\citet{cfcql} are repetitions of previously seen state-action pairs, compared to just $17\%$ in the data from \citet{formanek2023ogmarl}. This finding illuminates how complexities in multi-agent dynamics fail to be encapsulated solely in episode return values, and why we must be particularly careful.

\paragraph{\textbf{Takeaways.\;}} Much of the multi-agent literature uses datasets that are self-generated. Furthermore, there is often carelessness in the amount of effort put towards reproducibility and comparison. Datasets are often labelled with qualitative descriptors such as ``Good'', ``medium'' or ``bad'', descriptors which ought to have bearing on some property of the dataset. However, the most obvious property, episode return mean, is only sometimes reported, and rarely is episode return std reported. Our four progressive examples hope to illustrate the following key point. The mean, std, and distribution of episode returns are significantly important when learning from a multi-agent dataset, with a notable impact on learned performance. But they are not enough, and even when controlling for them, significant differences may still occur. Crucially, the outcomes from these examples do not hold in every scenario, but that is exactly the point: these peculiar dynamics can arise when not controlling for the dataset used, and what we witness can conceivably be leveraged by authors to erroneously present algorithmic advancements as more performant than what in reality is truly the case.

What is our main message, then? Ultimately, we see an urgent need for authors to pay attention to data in offline MARL research. At the very least, if authors are generating their own data for their experiments, they must provide easy access to this data for future work, hosted in perpetuity, with ample documentation of the generation procedure, the contents, and a reasonable quantitative description including the episode return distributions, the state-action coverage, and so on. In addition to such efforts, there is evidently a growing need for the standardisation of datasets in offline MARL. In the absence of a common foundation, it is impossible for researchers to speak the same language, introducing doubt into the veracity of algorithmic developments. We also see the groundwork of a fruitful new avenue of multi-agent research---how does the characteristics of a dataset impact multi-agent learning, with all the associated complex dynamics that inevitably arise?

\section{Putting Data at the Centre of Offline MARL} \label{sec:solutions}

Having illustrated how a lack of consideration for the impact of data in offline MARL is problematic, we take a step towards alleviating some of the issues we have identified in the previous section. By making three data-driven contributions to the community, we hope to bring data closer to the centre of research in the field. Our first contribution is a set of clear guidelines for how researchers should approach generating datasets in offline MARL. If there is an important reason for authors to generate their own datasets, there should at least be sound principles to follow. Secondly, we significantly enhance the standardisation of data in the field by converting over 80 datasets from prior works into a consistent format~\citep{flashbax}, which has an emphasis on speed, ease-of-use, clear documentation, and integration into existing frameworks. We upload these datasets to Hugging Face for reliable access in perpetuity.\footnote{\url{https://huggingface.co/datasets/InstaDeepAI/og-marl/}} We recommend using existing OG-MARL~\citep{formanek2023ogmarl} datasets for future research and encourage any new datasets generated by the community to be added to the repository following the standards and formats outlined there. That said, we still converted all the datasets we could get access to from prior work for the sake of continuity and the possibility of comparing with those works. Finally, we present an ever-growing repository of open-source tools that can be used to access, analyse, and edit these standardised datasets, for future research.

\subsection{Dataset guidelines} \label{sec:datasets-guidelines}
The gold standard solution is to standardise all of the existing datasets (which we attempt through OG-MARL) and to use a shared methodology when generating novel datasets. When generating a new dataset, there are certainly basic guiding principles that ought to be followed to ensure good scientific practice. We outline such guidelines in the blue box below.

\begin{figure}[ht]
    \begin{mybox}{Guidelines for generating new datasets for offline MARL research}{}
        \footnotesize
        \textbf{1. Is a new dataset really necessary?}
        \begin{itemize}
            \item Is there an existing dataset in the field you could use instead?
            \item If a new dataset is required for your research, make sure you document why and how exactly your dataset is different.
        \end{itemize}
        \textbf{2. Have you documented all of the relevant information regarding how you generated your data?}
        \begin{itemize}
            \item Document which environment you used and how people can access the environment. Make sure to be explicit about the version of the environment used. This is to ensure proper version control for comparisons as later environment versions might be released in the future (in some cases from authors different than the original creators). 
            \item Document relevant high-level environment properties e.g. number of agents, action size, observation size, sparse/dense reward and so on.
            \item Document how you generated the dataset. For example, which online MARL algorithm collected the experience and how did you sub-sample the data?
        \end{itemize}
        \textbf{3. Have you included a quantitative analysis of the composition of your data?}
        \begin{itemize}
            \item Report the following summary statistics for your datasets: episode returns min, mean, max and standard deviation, as well as the number of episodes and transitions in your dataset.
            \item Include plots of the episode return distribution e.g. histograms or violin plots.
            \item Include a measure of action-space coverage for your dataset e.g. Joint-SACo. 
        \end{itemize}
        \textbf{4. Can other researchers access your datasets? And will they still be able to in a year from now?}
        \begin{itemize}
            \item Make a download link easily accessible. Ensure the link will not expire and that downloads are successful from different regions of the world.
            \item Consider adding your datasets to a community-driven datasets repository such as OG-MARL~\citep{formanek2023ogmarl}.
            \item Use a dataset format that is widely adopted in the field. Or include sufficient documentation on how to load and use your dataset.
            \item Include a dataset licence.
        \end{itemize}
    \end{mybox}
    \label{fig:datasheet}
\end{figure}


\subsection{Standardising existing datasets} \label{sec:og-marl-datasets}
The single-agent offline RL community have benefited significantly from the widespread adoption of common datasets such as RL Unplugged~\citep{gulcehre2020rl} and D4RL~\citep{fu2020d4rl}. The offline MARL field could similarly benefit from the adoption of a common set of benchmark datasets.

Although there are multiple possible starting points for standardisation, we recommend OG-MARL~\citep{formanek2023ogmarl}. OG-MARL is solely focused on providing standardised datasets to the community, rather than hosting datasets that exist only as a by-product of algorithmic research. These datasets have already been cited in multiple works~\citep{formanek2023reduce, zhu2023madiff, yuan2023surveyprogresscooperativemultiagent, formanek2024dispelling, tilbury2024coordination, putla2024pilot, ruhdorfer2024overcookedgeneralisationchallenge, jing2024towards}. The repository is ever-growing and extendable by the community, and supports a wide range of environments.

We standardise the format of these datasets to the open-source and industry-supported utility of \texttt{Vault}~\citep{flashbax}, because of its focus on speed, clear documentation, and easy accessibility. For future offline MARL research, we strongly recommend using OG-MARL datasets in the Vault format. However, to enable comparisons to past works in the literature, we have additionally converted datasets from other authors into the Vault format. In Table~\ref{tab:vault_links}, we provide a list of the datasets converted and made available to the community in OG-MARL, hosted on the Hugging Face platform.


\begin{table}[t]
\centering
\scriptsize
\setstretch{1.3}
\caption{Complete list of datasets converted into the Vault format~\citep{flashbax}. Though we strongly recommend using OG-MARL in future research, we convert other datasets for continuity with past research. \emph{Environment sources:} MAMuJoCo~\citep{peng2021facmac}, SMACv1~\citep{samvelyan2019smac}, SMACv2~\citep{ellis2022smacv2}, MPE~\citep{lowe2017maddpg}, RWARE~\citep{christianos2020SharedExperience}}\label{tab:vault_links}
\vspace{1em}
\begin{adjustbox}{max width=\textwidth}
\begin{tabular}{l|lll}
\textbf{Source}& \textbf{Environment} & \textbf{Scenario} & \textbf{Datasets} \\ \hline
\cite{formanek2023ogmarl} & MAMuJoCo & \texttt{2halfcheetah} & \emph{good, medium, poor} \\
 & & \texttt{2ant} & \emph{good, medium, poor} \\
 & & \texttt{4ant} & \emph{good, medium, poor} \\
 & SMACv1 & \texttt{2s3z} & \emph{good, medium, poor} \\
 &  & \texttt{3m} & \emph{good, medium, poor} \\
 &  & \texttt{3s5z\_vs\_3s6z} & \emph{good, medium, poor} \\
 &  & \texttt{5m\_vs\_6m} & \emph{good, medium, poor} \\
 &  & \texttt{8m} & \emph{good, medium, poor} \\
 & SMACv2 & \texttt{terran\_5\_vs\_5} & \emph{replay} \\
 &  & \texttt{zerg\_5\_vs\_5} & \emph{replay} \\
\hline
\cite{omar} & MAMuJoCo & \texttt{2halfcheetah} & \emph{expert, medium-replay, medium, random} \\
 & MPE & \texttt{simple-spread} & \emph{expert, medium-replay, medium, random} \\
 & & \texttt{simple-tag} & \emph{expert, medium-replay, medium, random} \\
 & & \texttt{simple-world} & \emph{expert, medium-replay, medium, random} \\\hline
\cite{cfcql} & SMACv1 & \texttt{2s3z} & \emph{expert, medium-replay, medium, mixed} \\
 &  & \texttt{3s\_vs\_5z} & \emph{expert, medium-replay, medium, mixed} \\
 &  & \texttt{5m\_vs\_6m} & \emph{expert, medium-replay, medium, mixed} \\
 &  & \texttt{6h\_vs\_8z} & \emph{expert, medium-replay, medium, mixed} \\\hline
\cite{omiga} & SMACv1 & \texttt{corridor} & \emph{good, medium, poor} \\
 &  & \texttt{2c\_vs\_64zg} & \emph{good, medium, poor} \\
 &  & \texttt{5m\_vs\_6m} & \emph{good, medium, poor} \\
 &  & \texttt{6h\_vs\_8z} & \emph{good, medium, poor} \\
 & MAMuJoCo & \texttt{2ant} & \emph{expert, medium-expert, medium-replay, medium} \\
 & & \texttt{3hopper} & \emph{expert, medium-expert, medium-replay, medium} \\
 & & \texttt{6halfcheetah} & \emph{expert, medium-expert, medium-replay, medium} \\
 \hline
 \cite{alberdice} & RWARE & \texttt{tiny-2g} & \emph{expert} \\
 &  & \texttt{tiny-4g} & \emph{expert} \\
 &  & \texttt{tiny-6ag} & \emph{expert} \\
 &  & \texttt{small-2ag} & \emph{expert} \\
  &  & \texttt{small-4ag} & \emph{expert} \\
   &  & \texttt{small-6ag} & \emph{expert} \\\hline \hline
 & & \textbf{Total number of datasets:} & \textbf{88}
\\ 
\end{tabular}
\end{adjustbox}
\end{table}

\subsection{Dataset analysis tools} \label{sec:tools}

Our final contribution towards improving data awareness in offline MARL is a set of tools which can be used to download, subsample, combine, and analyse datasets. These tools, which live in OG-MARL, can be used on any dataset which conforms to the Vault API. The tools are accompanied by a demonstrative notebook\footnote{\url{https://github.com/instadeepai/og-marl/blob/main/examples/dataset_analysis_demo.ipynb}}, which explains how to use them and provides enough understanding of the Vault and OG-MARL systems to be able to work on custom tools and workflows. Our set of utilities are outlined below.

\begin{figure}[t]
    \begin{mybox}{Dataset utilities}{fig:analysis_tools}
        \footnotesize
        \textbf{Simplified loading of datasets}
        \begin{itemize}
            \item Support for downloading all 88 Vault datasets from OG-MARL.
            \item Support for downloading a Vault from a user-specified URL.
        \end{itemize}
        \textbf{Dataset analysis tools}
        \begin{itemize}
        \item Dataset structure: \texttt{describe\_structure} prints the pytree structure of each dataset in the Vault, and gives the number of transitions and trajectories in each dataset.
        \item Episode returns: \texttt{describe\_episode\_returns} plots for each dataset in the Vault the histogram and violin plot of its episode returns, and outputs a table containing the episode return mean, standard deviation, minimum, and maximum.
        \item Coverage properties: \texttt{describe\_coverage} produces a log-log plot of count frequencies of unique state-action pairs, as well as the Joint-SACo value for each dataset.
        \item Summary: \texttt{descriptive\_summary} plots episode return histograms, and outputs a table containing episode return mean, standard deviation, minimum and maximum, Joint-SACo, number of transitions and number of trajectories in each dataset.
    \end{itemize}
    \textbf{Tools to subsample and combine Vaults}
        \begin{itemize}
            \item Subsample a Vault to within one trajectory's length of a specified number of transitions.
            \item Combine a list of datasets into one larger dataset.
            \item Subsample two datasets to have near-identical episode return distributions.
            \item Subsample a dataset to have a specific episode return distribution.
        \end{itemize}
    \end{mybox}
\end{figure}


We provide a demonstration of our tools using the \texttt{2s3z} scenario from SMACv1~\citep{samvelyan2019smac}, with the dataset from OG-MARL~\citep{formanek2023ogmarl}. We focus on dataset analysis (where we give insights into dataset composition), as well as subsampling and combining tools (which can be used for a variety of reasons: to make datasets smaller so that they use less memory, to create datasets for ablations, and to combine datasets when more training data is required).

\begin{figure}
    \centering
    \begin{subfigure}[p]{\textwidth}
        \centering
        \caption{Tabular values returned from the \texttt{descriptive\_summary} function}
        \adjustbox{width=0.8\textwidth}{
        \begin{tabular}{lccccccc}
            \hline
            Dataset & Mean & Stddev & Min & Max & \# Transitions & \# Trajectories & Joint-SACo \\
            \hline
            Good & 18.32 & 2.95 & 0.00 & 21.62 & 995\,829 & 18\,616 & 0.98 \\
            Medium & 12.57 & 3.14 & 0.00 & 21.30 & 996\,256 & 18\,605 & 0.98 \\
            Poor & 6.88 & 2.06 & 0.00 & 13.61 & 996\,418 & 9\,942 & 0.96 \\
            \hline
            \hline
        \end{tabular}
        }
        \label{fig:enter-label}
    \end{subfigure}
    \\\vspace{1em}
    \begin{subfigure}[p]{\textwidth}
        \centering
        \includegraphics[width=0.7\linewidth]{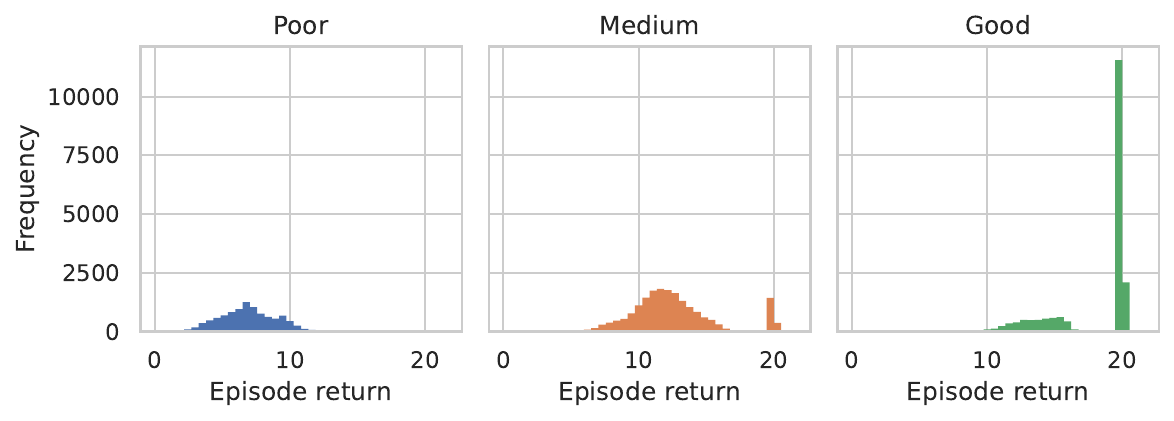}
        \caption{Histograms returned from the \texttt{descriptive\_summary} function}
        \label{fig:all_hist}
    \end{subfigure}
    \caption{The results returned when calling \texttt{descriptive\_summary} on the \texttt{2s3z} Vault from OG-MARL~\citep{formanek2023ogmarl}}
    \label{fig:output_descriptive_summary}
\end{figure}

\paragraph{Analysis}
Our analysis tools cover all requirements stipulated in the analysis section of our dataset generation guidelines. We provide four high-level functions to generate various insights for a user-specified selection of datasets in the Vault format. Calling \texttt{descriptive\_summary} with a provided Vault will generate a summary such as the one illustrated by Figure~\ref{fig:output_descriptive_summary}, with both tabular and histogram information returned. Users can further access episode return violin plots by calling \texttt{describe\_episode\_returns}, detailed structural information about the Vault by calling \texttt{describe\_structure}, and state-action count information by calling \texttt{describe\_coverage}.
 
From these outputs, we can now analyse the dataset and notice interesting insights. For example, we can see in Table \ref{fig:output_descriptive_summary} that the \texttt{Poor} dataset contains far fewer trajectories than the \texttt{Good} and \texttt{Medium} datasets, despite containing a similar number of transitions. On average, the episodes in the \texttt{Poor} dataset contain more than 100 transitions, which means that the episodes usually roll out to their full length. We also notice that there is not much of a difference between the Joint-SACo of the datasets, though each dataset's coverage is diverse, with at most $4\%$ being repeated pairs in each case. The information gained here can at a glance help a researcher understand a dataset better.

\begin{figure}[t]
    \centering
    \begin{subfigure}[t]{0.45\textwidth}
        \centering
        \includegraphics[width = \linewidth]{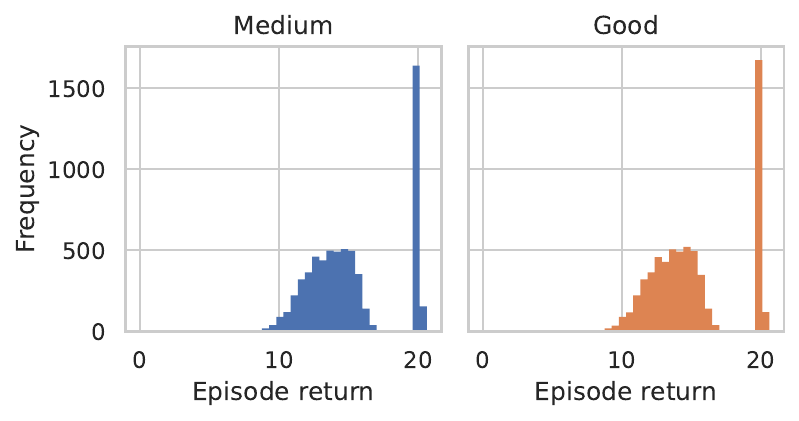}
        \caption{Original \texttt{Medium} and \texttt{Good} datasets subsampled to have similar episode return distributions.}
    \end{subfigure} \\
    \begin{subfigure}[p]{0.45\textwidth}
         \centering
         \includegraphics[width=0.7\textwidth]{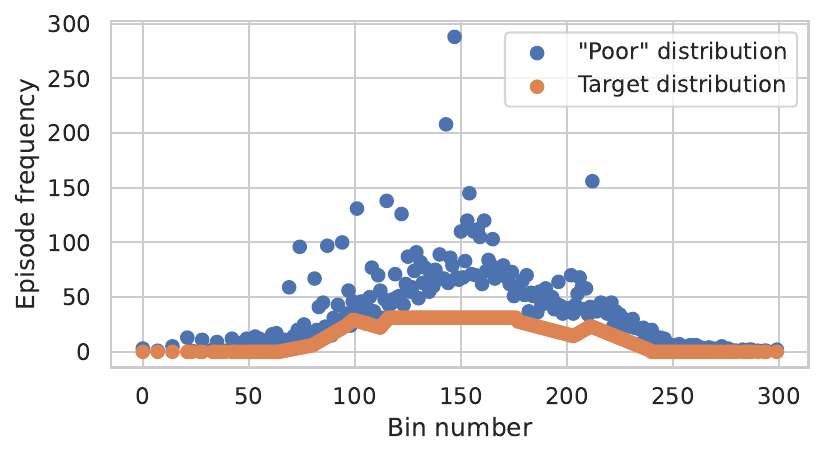}
         \caption{\texttt{Poor} dataset distribution versus a target distribution.}
         \label{fig:tbl_mtn_pdf}
    \end{subfigure}
    \hspace{2em}
    \begin{subfigure}[p]{0.45\textwidth}
        \centering
         \includegraphics[width=0.7\textwidth]{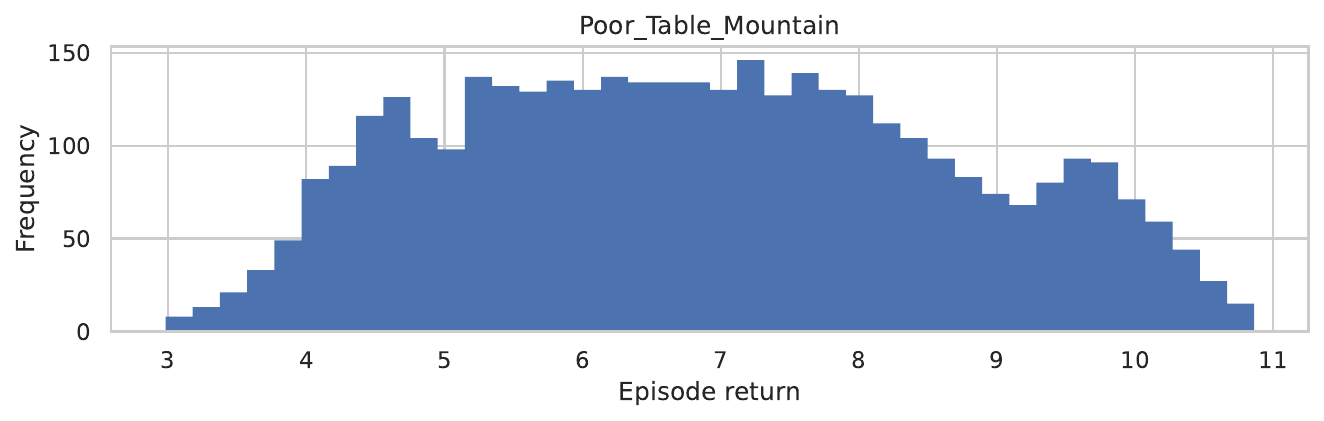}
         \caption{Histogram of the dataset with the desired target distribution, after being subsampled from \texttt{Poor}.}
            \label{fig:tbl_mtn_hist}
    \end{subfigure} \\
    \begin{subfigure}[t]{0.7\textwidth}
        \centering
        \includegraphics[width = \linewidth]{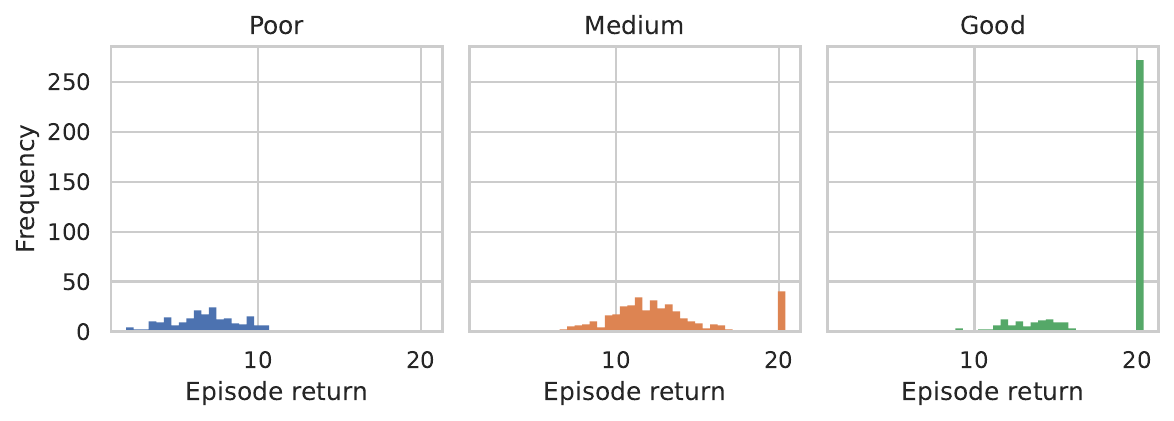}
        \caption{Datasets subsampled to 20\,000 transitions each.}        
    \end{subfigure}\hfill
    \begin{subfigure}[t]{0.25\textwidth}
        \centering
        \includegraphics[width = \linewidth]{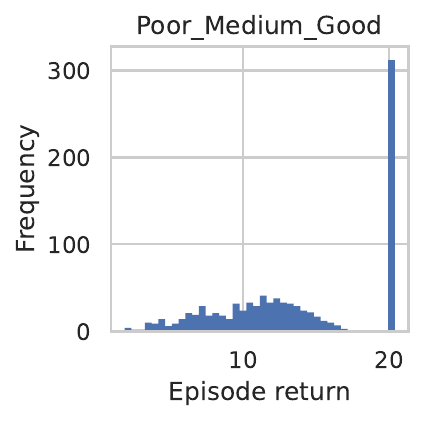}
        \caption{Combined datasets.}        
    \end{subfigure}
        
    
    \caption{Illustrative dataset subsampling examples, using the \texttt{2s3z} Vault.}
    \label{fig:dataset_subsampling}
\end{figure}

\paragraph{Subsampling and combining datasets}
Our tools also allow researchers to stitch and slice datasets. While it is important to keep datasets standardised, experiments may require datasets which do not yet exist; there may be constrained memory requirements, new experiments, or ablations requiring new datasets. Rather than expecting researchers to create entirely new datasets for such experiments, which might introduce inconsistencies between datasets available in the field, we give researchers the tools to flexibly subsample and combine existing datasets according to the properties that their experiments require. Figure \ref{fig:dataset_subsampling} illustrates an assortment of such examples---showing how datasets can be subsampled or combined, and showing how desired distributions can be systematically created. We also provide a notebook in the OG-MARL open-source repository that demonstrates how to subsample a dataset according to a user-specified episode return histogram.\footnote{\url{https://github.com/instadeepai/og-marl/blob/main/examples/dataset_subsampling_demo.ipynb}} Our hope is that even though our subsampling tools allow researchers to manipulate data more easily, our analysis tools make it easy enough to uncover dataset discrepancies that researchers should be dissuaded from changing datasets without disclosing their amendments.

Our work opens up an easy-to-use pipeline for data use, generation and understanding in offline MARL. Researchers can now easily expose the contents of datasets for flexible and well-informed access to the material on which their algorithms train. We hope the impact of our work is better understanding not only of datasets in offline MARL, but also of the environments we use and how a data collecting policy interacts with these.

\section{Conclusion} \label{sec:conclusion}
In this paper, we have surveyed the literature and found that data is largely neglected in the data-driven field of offline MARL. We have shown why paying attention to data in offline MARL research is crucial, through simple yet illuminating examples. We contribute to the community: a guideline for generating multi-agent datasets; a standardised repository of over 80 \texttt{environment-scenario-quality} combinations, with a well-documented and accessible API; and useful tooling to aid the understanding of these datasets. In conjunction, these efforts aim to catalyse progress by aligning the field towards scientific rigour. In doing so, along with other standardisation efforts---e.g. standardised baselines~\citep{formanek2024dispelling}---we feel that the discipline is ripe with opportunity to solve hard problems. We encourage researchers to adopt and extend our offerings, working collectively to push the field forward. By working from a strong foundation together, significant breakthroughs can be made.

\newpage
\impact{
Our contribution opens the door to an offline MARL field in which careful attention is paid to data. However, the path from our contribution to that goal needs to be taken collectively by the community. If our work becomes widely adopted, then we may be shaping what datasets in offline MARL research look like - standardisation by nature places some restriction on the object being standardised, but our guidelines and contributions take into account that some level of flexibility is also required for progress. It is possible that, using our tools, the field could trend towards either using pre-existing standardised datasets (which is preferable) or again generating their own datasets for new experiments (which will be easy to do using our tooling). It is also true that the subsampling and combining tools which we provide can both improve accessibility and standardisation of datasets but also allow researchers to manipulate data without disclosing their alterations.


We must consider not only research impact, but real-world impact. Our long-term goal for impact is to build solid foundations on which offline MARL can develop, which may accelerate the progress in the field. The potential effects of progress in offline MARL are vast, but our work specifically assists in standardising and analysing datasets. We hope that transparency around the nature of datasets will filter through not only in research but also into real-world scenarios. Dataset transparency in the real world is, however, a far more complex topic since datasets from real-world scenarios may contain sensitive information. Care must therefore be taken in performing and presenting our suggested analyses on real-world datasets.

As with any Deep Learning research, our contribution is likely to have computational expense implications. If datasets are standardised, fewer new datasets need to be created, and fewer baselines need to be rerun. Both of these effects reduce computational expense. Overall, while we have hopes for the impact of our research, we acknowledge areas in which it may be misused. We urge the community to adopt the guidelines of our work in the hopes that progress can be made with care.}





\acks{This work was supported by InstaDeep Ltd.}

\newpage
\vskip 0.2in
\bibliography{main}

\end{document}